\documentclass[preprint,12pt]{elsarticle}

\usepackage{amssymb}
\usepackage{amsmath}

\usepackage{makecell}
\usepackage{booktabs}
\usepackage{multirow}
\usepackage{algpseudocode}
\usepackage{algorithm}
\usepackage{lscape} 
\usepackage{url} 

\algnewcommand{\IfThenElse}[3]{
  \State \algorithmicif\ #1\ \algorithmicthen\ #2\ \algorithmicelse\ #3}

\journal{}

\begin{document}

\begin{frontmatter}



\title{Lung-DDPM+: Efficient Thoracic CT Image Synthesis using Diffusion Probabilistic Model} 


\author[aff_crchu,aff_dom,aff_crc,aff_iid]{Yifan Jiang}
\author[aff_ui]{Ahmad Shariftabrizi}
\author[aff_crchu,aff_dom,aff_crc,aff_iid]{Venkata SK. Manem \corref{cor1}}

\cortext[cor1]{Corresponding author: venkata.manem@crchudequebec.ulaval.ca}

\affiliation[aff_crchu]{organization={Centre de recherche du CHU de Québec-Université Laval},
            addressline={2260 boul. Henri-Bourassa},
            city={Québec},
            postcode={G1J 0J9},
            state={QC},
            country={Canada}}
\affiliation[aff_dom]{organization={Department of Molecular Biology, Medical Biochemistry and Pathology, Université Laval},
            addressline={Ferdinand Vandry Pavillon, 1050 Rue de la Médecine},
            city={Québec},
            postcode={G1V 0A6},
            state={QC},
            country={Canada}}
\affiliation[aff_crc]{organization={Cancer Research Center, Université Laval},
            addressline={9 Rue McMahon},
            city={Québec},
            postcode={G1R 3S3},
            state={QC},
            country={Canada}}
\affiliation[aff_iid]{organization={Big Data Research Center, Université Laval},
            addressline={Adrien Pouliot Pavilion, 1065 Av. de la Médecine},
            city={Québec},
            postcode={G1V 0A6},
            state={QC},
            country={Canada}}
\affiliation[aff_ui]{organization={Department of Radiology, Carver College of Medicine - The University of Iowa},
            addressline={200 Hawkins Drive},
            city={Iowa City},
            postcode={52242},
            state={IA},
            country={United States}}


\begin{abstract}
Generative artificial intelligence (AI) has been playing an important role in various domains. Leveraging its high capability to generate high-fidelity and diverse synthetic data, generative AI is widely applied in diagnostic tasks, such as lung cancer diagnosis using computed tomography (CT). However, existing generative models for lung cancer diagnosis suffer from low efficiency and anatomical imprecision, which limit their clinical applicability. To address these drawbacks, we propose Lung-DDPM+, an improved version of our previous model, Lung-DDPM. This novel approach is a denoising diffusion probabilistic model (DDPM) guided by nodule semantic layouts and accelerated by a pulmonary DPM-solver, enabling the method to focus on lesion areas while achieving a better trade-off between sampling efficiency and quality. Evaluation results on the public LIDC-IDRI dataset suggest that the proposed method achieves $8\times$ fewer FLOPs (floating point operations per second), $6.8\times$ lower GPU memory consumption, and $14\times$ faster sampling compared to Lung-DDPM. Moreover, it maintains comparable sample quality to both Lung-DDPM and other state-of-the-art (SOTA) generative models in two downstream segmentation tasks. We also conducted a Visual Turing Test by an experienced radiologist, showing the advanced quality and fidelity of synthetic samples generated by the proposed method. These experimental results demonstrate that Lung-DDPM+ can effectively generate high-quality thoracic CT images with lung nodules, highlighting its potential for broader applications, such as general tumor synthesis and lesion generation in medical imaging. The code and pretrained models are available at \url{https://github.com/Manem-Lab/Lung-DDPM-PLUS}.
\end{abstract}




\end{frontmatter}



\section{Introduction}
\label{sec1}
Lung cancer is one of the greatest threats to human well-being. In 2022, it remained the most frequently diagnosed cancer worldwide, with approximately 2.5 million new cases, accounting for 12.4\% of all cancer diagnoses. Furthermore, lung cancer continues to be the leading cause of cancer-related mortality, responsible for approximately 1.8 million deaths (18.7\% of total cancer deaths) in 2022 \cite{bray2024global}. For small, localized tumors (stage I), surgical resection of non-small cell lung cancer (NSCLC) offers a favorable prognosis, with 5-year survival rates of up to 70\%. In contrast, cases diagnosed with distant metastatic disease (stage IV) have a 1-year survival rate of only 15-19\%, compared to 81-85\% for stage I \cite{blandin2017progress}. These statistics underscore the critical importance of early detection of lung nodules and intervention in lung cancer diagnosis.

Given the rapid advancement of AI in healthcare, numerous deep learning-based models have been deployed across various domains. However, this progress exacerbates data scarcity, which significantly impacts AI-assisted diagnostic models for early lung cancer detection. Recently, generative AI has garnered increasing attention for its ability to provide high-quality synthetic samples, enhancing model performance and mitigating data limitations. In radiological imaging, generative AI applications extend to X-ray \cite{salehinejad2018synthesizing}, computed tomography (CT) \cite{jiang2020covid}, positron emission tomography (PET) \cite{luo2022adaptive}, magnetic resonance imaging (MRI) \cite{dorjsembe2024conditional}, and ultrasound \cite{liang2022sketch}. Among these, CT image synthesis remains one of the most challenging tasks due to its high computational complexity and the intricate texture details of anatomical structures.

With the emergence of generative AI, various models for CT image synthesis have been introduced over the past decade. Since 2020, our team has focused on generative models for thoracic CT synthesis, initially proposing a generative adversarial network (GAN)-based approach for 2D COVID-19 CT image synthesis \cite{jiang2020covid}. More recently, we advanced this work by introducing Lung-DDPM \cite{jiang2025lungddpm}, a denoising diffusion probabilistic model (DDPM) for 3D CT image synthesis with lung nodules. While Lung-DDPM addressed three key limitations of existing methods—anatomical imprecision, imperfect blending, and inconsistency—SOTA models still suffer from low efficiency in generating 3D thoracic CT images.

\begin{figure}[!ht]
\centering
\includegraphics[width=13cm]{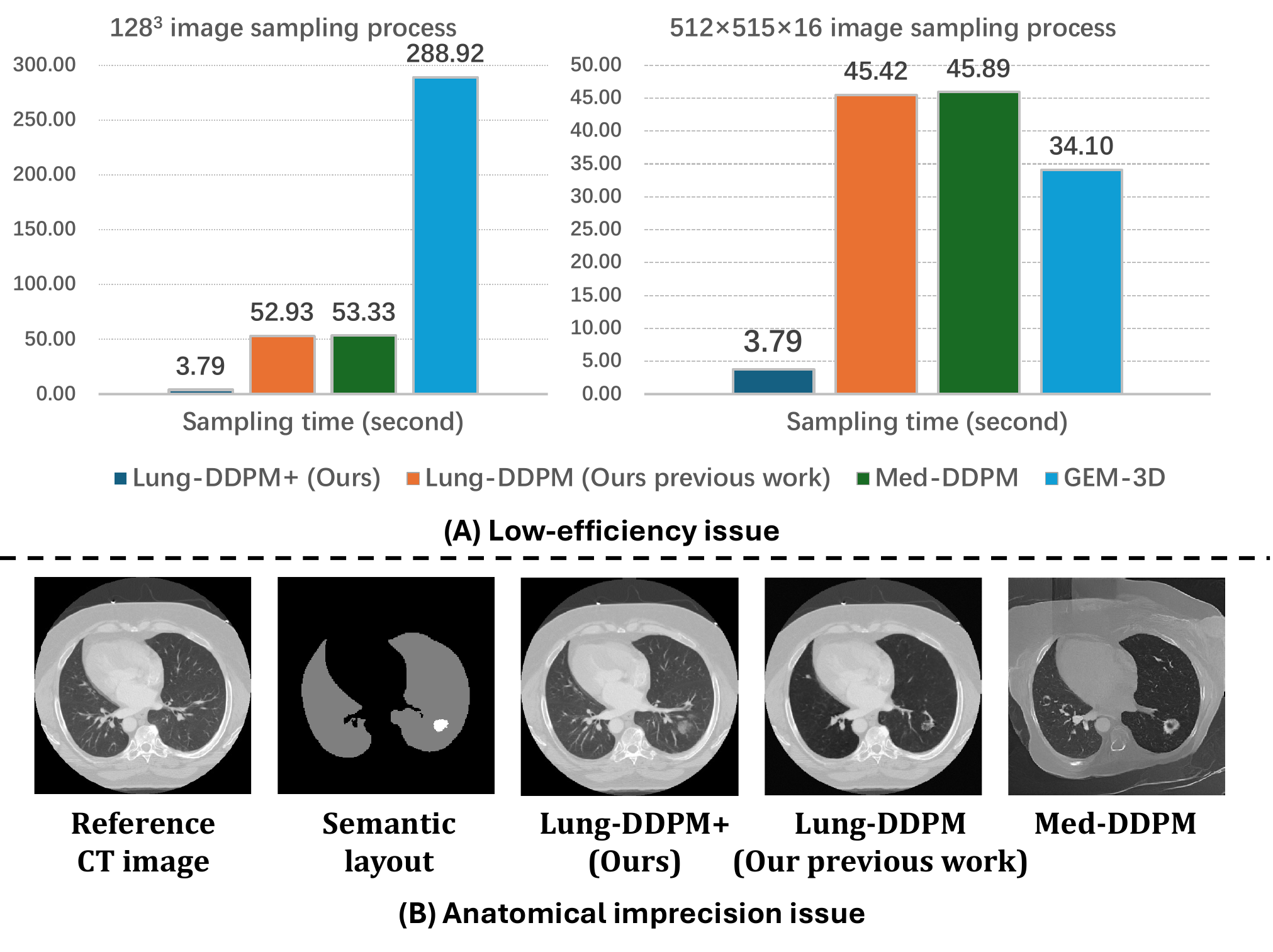}
\caption{Demonstration of low efficiency and anatomical imprecision issues in SOTA generative models for thoracic CT image synthesis.}
\label{fig1} 
\end{figure}

As illustrated in Figure \ref{fig1} (A), we compare the sampling times of the proposed Lung-DDPM+ with three SOTA models—Lung-DDPM \cite{jiang2025lungddpm}, Med-DDPM \cite{dorjsembe2024conditional}, and GEM-3D \cite{zhu2024generative}—for generating thoracic CT images of two different sizes. To synthesize a single CT image, SOTA generative models require anywhere from tens to over two hundred seconds, resulting in low efficiency during the sampling process. This inefficiency severely limits the practical deployment of generative models in downstream medical applications, particularly in time-sensitive and cost-sensitive scenarios.

Beyond the low efficiency issue, we also observed that SOTA generative models tend to synthesize imprecise anatomical structures when no semantic layout guidance is available. In Figure \ref{fig1} (B), we illustrate a case generated by Lung-DDPM and Med-DDPM, both exhibiting anatomical imprecision. Compared to the proposed Lung-DDPM+ sample, the synthetic image produced by Med-DDPM failed to maintain anatomical precision in extra-pulmonary areas. While Lung-DDPM preserved extra-pulmonary regions, its synthetic sample lacked pulmonary vessel and bronchus details. In clinical practice, experienced radiologists manually annotate detailed semantic layouts for organs and lesions. However, this process is time-consuming and costly \cite{ma2021toward}, making well-labeled medical imaging datasets scarce for researchers. Therefore, developing a feasible approach to address the anatomical imprecision issue in scenarios with limited annotations remains a critical research challenge.

To address the low efficiency and anatomical imprecision of current generative models in thoracic CT image synthesis, we propose Lung-DDPM+, an improved denoising diffusion probabilistic model (DDPM) based on Lung-DDPM. By leveraging the novel pulmonary DPM-solver, our approach generates various-sized lung nodules instead of entire pulmonary regions and seamlessly embeds them into healthy cases, preserving most anatomical details. Furthermore, the pulmonary DPM-solver reduces the synthetic image size from $128^3$ to $64^3$, significantly lowering resolution compared to other SOTA models and playing a crucial role in accelerating the sampling process. We summarize the key contributions of this paper as follows:

\begin{itemize} 
\item [(1)] We propose a novel pulmonary DPM-solver to address the low efficiency issue that hinders the widespread adoption of generative models in thoracic image synthesis. This DPM-solver significantly reduces reliance on high-performance hardware and accelerates the sampling process.
\item [(2)] Building upon Lung-DDPM, the proposed Lung-DDPM+ further mitigates the anatomical imprecision issue by directly generating lung nodules. This approach enhances the preservation of anatomical structures beyond synthetic lung nodules, including pulmonary vessels and the bronchial tree.
\item [(3)] Our method has low data requirements and does not rely on externally sourced healthy cases, making it applicable to datasets in which all cases contain nodules (e.g., LIDC-IDRI \cite{samuel2011lung}). By synthesizing additional nodule-positive samples with greater diversity, our approach enhances the robustness and performance of downstream segmentation models. While healthy cases are not required, they can be incorporated to further expand the diversity of the synthetic dataset.
\item [(4)] Lung-DDPM+ surpasses other SOTA generative models in model efficiency evaluation and two downstream lung nodule segmentation tasks, demonstrating its superior performance across various applications.
\item [(5)] This effective thoracic CT image synthesis method substantially reduces the need for high-quality labeled data. Unlike Lung-DDPM, which requires radiologists to provide nodule segmentation, the proposed method automatically generates both nodule segmentation and the corresponding CT image within a single workflow.
\item [(6)] The high-quality synthetic CT images generated by Lung-DDPM+ have the potential to alleviate data scarcity in AI-assisted clinical applications, such as lung cancer screening, survival rate prediction, and therapy planning.
\end{itemize}

\section{Related works}
\label{sec2}
\subsection{Efficient denoising diffusion probabilistic models}
Since the introduction of DDPM \cite{ho2020denoising}, it has emerged as one of the most advanced generative models due to its strong ability to generate high-quality data across various applications. However, low efficiency remains a critical limitation, restricting both its performance and deployment. To address this issue, researchers have explored multiple strategies to enhance DDPM efficiency. Improved model architectures, such as latent space diffusion models \cite{rombach2022high}, allow DDPMs to operate in a compressed space, significantly reducing computational demands. Adaptive noise scheduling methods, including PriorGrad \cite{lee2021priorgrad}, introduce data-dependent priors to accelerate convergence and improve sample quality. Additionally, hybrid model integration, which combines diffusion models with pre-trained generative architectures like GANs and VAEs \cite{dhariwal2021diffusion}, leverages the strengths of both approaches to enhance sampling speed. Further optimizations focus on the sampling process, where techniques such as efficient sampling schedules \cite{watson2021learning} and early stopping mechanisms \cite{lyu2022accelerating} reduce the number of denoising steps while maintaining generation quality. Despite these advancements, research on optimizing DDPM sampling efficiency in medical imaging remains limited, necessitating further exploration to improve its feasibility in clinical applications.

\subsection{Generative models for medical imaging}
Over the past decade, generative models have been extensively utilized in medical imaging, driven by rapid advancements in deep learning. For instance, Jiang et al. \cite{jiang2020covid} employed a conditional generative adversarial network (cGAN) to synthesize COVID-19 CT images, facilitating data augmentation for improved diagnostic models. Dorjsembe et al. \cite{dorjsembe2024conditional} introduced conditional diffusion models for semantic 3D brain MRI synthesis, enhancing the generation of realistic brain images for various clinical applications. Zhu et al. \cite{zhu2024generative} proposed a generative enhancement approach for 3D medical images, focusing on improving image quality and resolution. Diffusion models have also gained traction in medical imaging, as highlighted in a recent review by Hein et al. \cite{hein2024physics}, which explores the transformative role of physics-inspired generative models in this domain. Among these advancements, Jiang et al. \cite{jiang2025lungddpm} developed Lung-DDPM, a semantic layout-guided diffusion model for lung CT image synthesis, demonstrating superior performance over other state-of-the-art (SOTA) methods. However, like other approaches, it suffers from low efficiency, a key challenge this paper aims to address.

\section{Proposed method}
\label{sec3}
\subsection{Overview}

\begin{figure}[!ht]
\centering
\includegraphics[width=13cm]{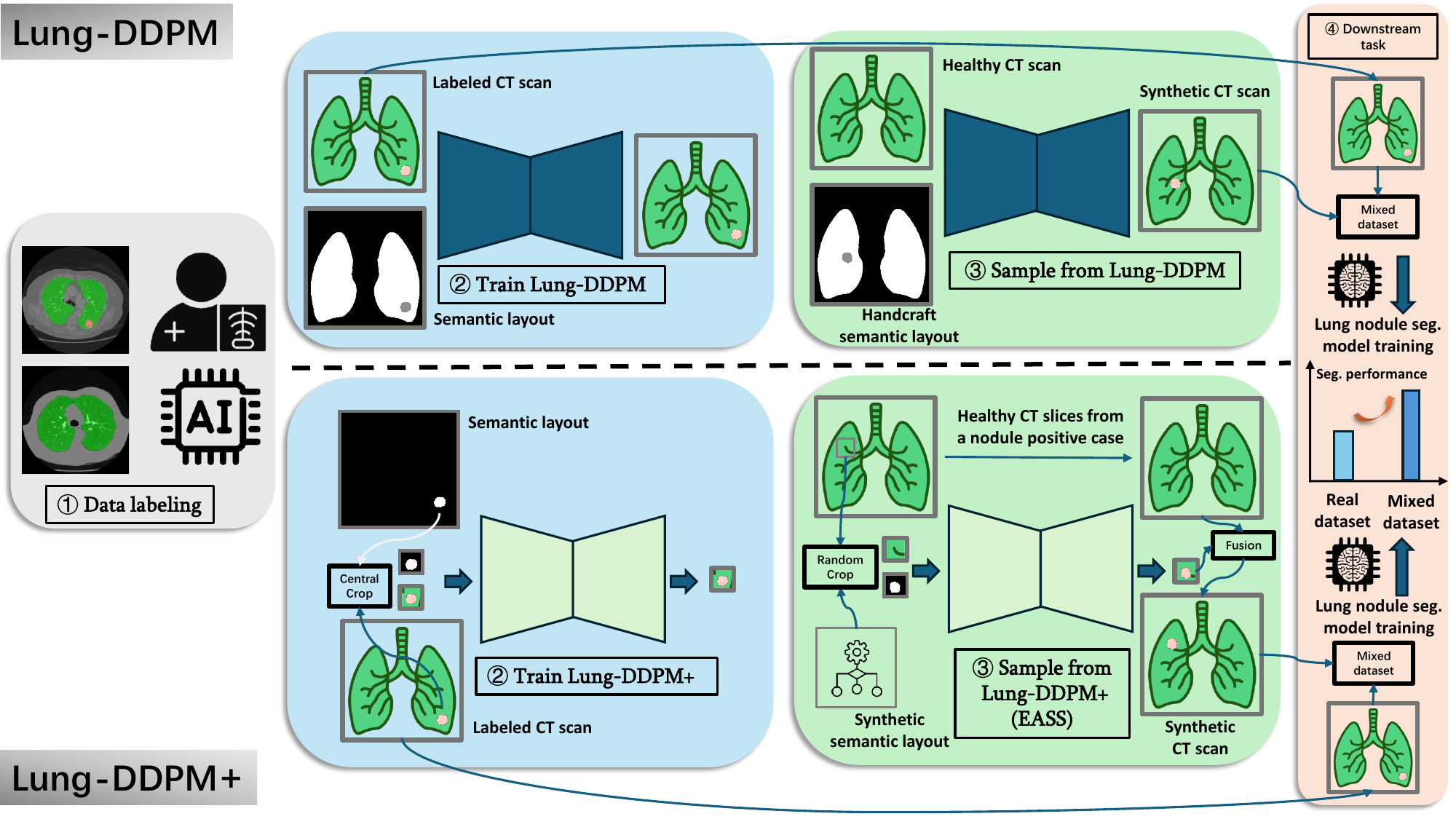}
\caption{Comparasion of workflows for Lung-DDPM and Lung-DDPM+. The workflow for both models consists of four distinct steps: \textcircled{\raisebox{-0.9pt}{1}} Data labeling, \textcircled{\raisebox{-0.9pt}{2}} Model training, \textcircled{\raisebox{-0.9pt}{3}} Sampling from the model, and \textcircled{\raisebox{-0.9pt}{4}} Downstream task.}
\label{fig2} 
\end{figure}

In Figure \ref{fig2}, we present an overview of Lung-DDPM+'s workflow and compare it with its predecessor, Lung-DDPM. The first step \textcircled{\raisebox{-0.9pt}{1}} in both methods begins with data labeling. CT scans collected from medical institutions are evaluated and labeled by experienced radiologists for lung nodules, while pulmonary regions are automatically annotated using AI-assisted diagnostic systems. During the model training step \textcircled{\raisebox{-0.9pt}{2}}, instead of using the entire pulmonary region as in Lung-DDPM, we train Lung-DDPM+ using centrally cropped lung nodule scans, significantly reducing model scale and training time. In the model sampling stage \textcircled{\raisebox{-0.9pt}{3}}, referred to as the effective anatomically aware sampling (EAAS) process, the proposed EAAS takes a randomly cropped pulmonary region and generates a realistic lung nodule based on it. This synthetic lung nodule is then integrated into the original CT scan through a fusion process, forming the final synthetic sample. During the downstream task \textcircled{\raisebox{-0.9pt}{4}}, a real-synthetic mixed dataset is created by combining both synthetic and real samples. This mixed dataset serves as the training set for the lung nodule segmentation task, where an improvement in segmentation performance is expected compared to training with real samples alone.

\subsection{Pulmonary DPM-solver}
Unlike Lung-DDPM, which operates on the entire pulmonary region, the proposed method focuses on the effective synthesis of nodule regions. To accelerate the anatomically aware sampling (AAS) process, we introduce a pulmonary DPM-solver, which extends the original DPM-solver \cite{lu2022dpm,lu2022dpm++} by dynamically blending nodules with surrounding pulmonary areas during the EAAS sampling process.

Diffusion Probabilistic Models (DPMs) have emerged as a state-of-the-art generative modeling approach, demonstrating remarkable performance in high-dimensional structured data generation. These models learn to reverse a predefined forward diffusion process, where Gaussian noise is progressively added to an input sample until it becomes a random distribution. Sampling from the trained model involves solving the corresponding reverse stochastic differential equation (SDE) or its deterministic probability flow ordinary differential equation (ODE):

\begin{equation}
d\mathbf{x} = \left[ f(t)\mathbf{x} - \frac{1}{2}g^2(t)\nabla_{\mathbf{x}} \log p_t(\mathbf{x}) \right] dt
\label{eq1}
\end{equation}
where \(\mathbf{x}\) represents the input data, $f(t)$ is the drift coefficient at time $t$, $g(t)$ is the diffusion coefficient at $t$, and \(\nabla_{\mathbf{x}} \log p_t(\mathbf{x})\) is the learned score function (the gradient of the log probability density at time $t$).

Despite their impressive generative capabilities, standard DPMs suffer from slow inference, often requiring hundreds or thousands of function evaluations (NFEs) for high-quality sampling. To mitigate this limitation, DPM-Solver \cite{lu2022dpm} was introduced as a fast, high-order ODE solver that significantly accelerates sampling while preserving sample fidelity.

DPM-Solver is a deterministic numerical method designed to efficiently solve the Probability Flow ODE using high-order approximation schemes. It employs an adaptive multi-step solver, leveraging second- and third-order methods to approximate the reverse diffusion trajectory with minimal NFEs. The general formulation follows:

\begin{equation}
\mathbf{x}(t + \Delta t) = \mathbf{x}(t) + \Delta t F_1 + \frac{(\Delta t)^2}{2} F_2 + \frac{(\Delta t)^3}{6} F_3
\label{eq2}
\end{equation}
where $\Delta t$ is the timestep, and \( F_1 \), \( F_2 \), and \( F_3 \) are progressively computed derivatives of the score function. This approach reduces sampling complexity from thousands of steps to 10--20 steps, making DPMs significantly more efficient.

Building upon DPM-Solver, DPM-Solver++ \cite{lu2022dpm++} further optimizes inference speed and quality by integrating both stochastic and deterministic solvers in a hybrid framework. In early diffusion steps, stochastic updates enhance sample diversity, while deterministic high-order solvers refine results in later timesteps:

\begin{equation}
\mathbf{x}(t + \Delta t) = \mathbf{x}(t) + \Delta t F_1 + \frac{(\Delta t)^2}{2} F_2 + \gamma g(t) \Delta \mathbf{w}
\label{eq3}
\end{equation}
where \(\gamma\) controls the level of stochasticity, ensuring a balance between efficiency and generative quality, and $\Delta \mathbf{w}$ is a Wiener process increment (i.e., standard Gaussian noise scaled by the time step).

DPM-Solver++ reduces NFEs to as few as 5–10, making diffusion-based generation feasible for real-time applications. However, in medical imaging tasks, where structural consistency and conditional control are crucial, further improvements are necessary. While DPM-Solver++ optimizes speed and efficiency, it does not explicitly incorporate domain-specific conditions, which are critical for medical applications such as lung nodule analysis. In thoracic CT image synthesis, lung nodules exhibit complex spatial structures that require conditioned diffusion processes to ensure clinically meaningful outputs.

To address this gap, we introduce Pulmonary DPM-Solver, a modified version of DPM-Solver++ designed for lung nodule segmentation-aware generation. Our approach integrates structural prior knowledge by masking the noise predictor input with a segmentation condition tensor. This modification ensures that the generated samples are spatially and structurally constrained, significantly improving the fidelity of medical image synthesis.

In conventional DPM-Solver++, the noise predictor function operates solely on the corrupted input \(\mathbf{x}_t\). We modify this formulation by explicitly incorporating the lung nodule segmentation map \(\mathbf{c}\), where \(\mathbf{c}\) is a binary lung nodule segmentation mask:

\begin{equation}
\mathbf{x}_{t+1} = \text{DPM-Solver++}(\mathbf{x}_t, \mathbf{c})
\label{eq4}
\end{equation}

Instead of predicting noise solely from \(\mathbf{x}_t\), our modified score function is conditioned as:

\begin{equation}
\epsilon_\theta(\mathbf{x}_t, t, \mathbf{c})
\label{eq5}
\end{equation}
where \(\epsilon_\theta\) is the neural network estimating the noise residual at timestep \(t\), conditioned on both the current sample \(\mathbf{x}_t\) and segmentation tensor \(\mathbf{c}\). This conditioned diffusion process enforces anatomical constraints, ensuring that the generated CT samples maintain realistic lung and nodule structures.

Pulmonary DPM-Solver builds upon the fast sampling efficiency of DPM-Solver++ by introducing segmentation-conditioned diffusion for thoracic CT image synthesis. By incorporating anatomical priors via segmentation-conditioned noise prediction, our method ensures structural consistency, reduces sampling complexity, and enhances the realism of generated medical images.

\subsection{Effective anatomically aware sampling (EAAS)}
\begin{figure}[!ht]
\centering
\includegraphics[width=14cm]{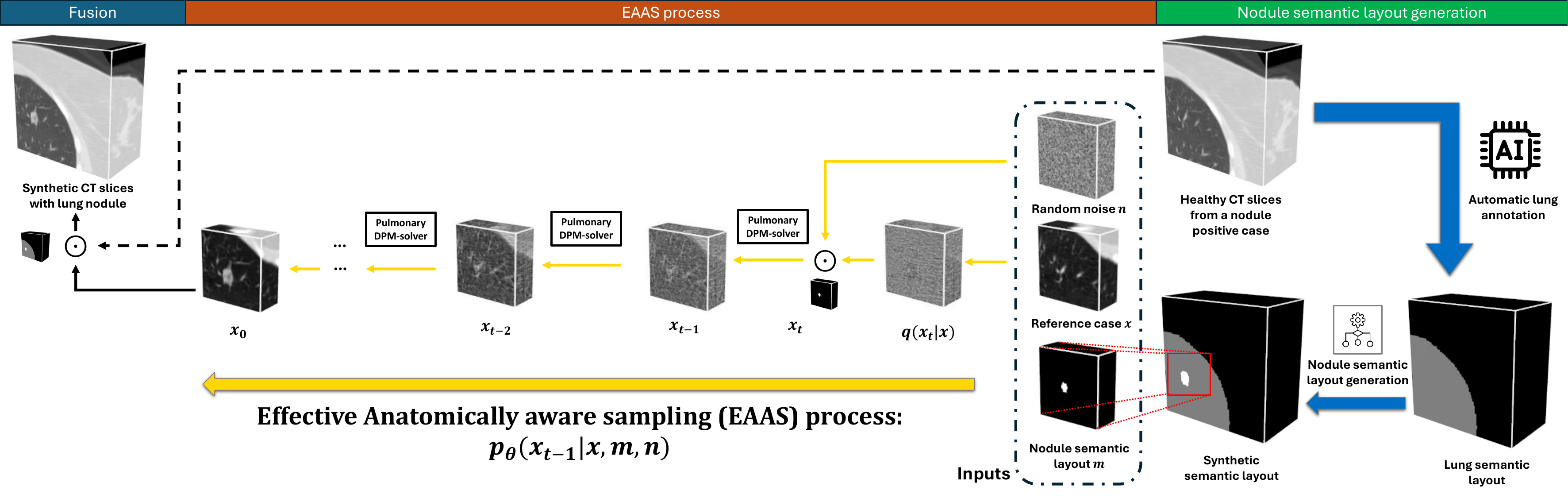}
\caption{The sampling process of the proposed method consists of three key steps: (1) Nodule semantic layout generation, (2) Effective anatomically aware sampling (EAAS) process, and (3) Fusion.}
\label{fig3} 
\end{figure}

\subsubsection{Nodule semantic layout generation}
The generation of pulmonary CT images relies heavily on high-quality lung nodule semantic layouts, which are typically created by experienced radiologists in the Lung-DDPM workflow. To reduce radiologists' workload and enhance the efficiency of the generation process, we introduce a rule-based nodule semantic layout generation method. Specifically, following \cite{chen2024towards} and \cite{hu2023label}, we first generate realistic tumor-like shapes using ellipsoids that match the size distribution of our dataset. Unlike \cite{chen2024towards} and \cite{hu2023label}, our approach randomly samples healthy CT slices from an existing nodule-positive case rather than from an entirely novel healthy case. This strategy significantly reduces data dependency, particularly in data scarcity scenarios. For the downstream 2D lung nodule segmentation task, we randomly crop a $16 \times 512 \times 512$ patch from a nodule-positive CT image, ensuring that the selected slices do not contain lung nodules. We then generate a nodule mask within the automatically annotated lung regions. For the downstream 3D lung nodule segmentation task, we follow the same procedure as in the 2D task, but instead use a $128^3$ patch.

\subsubsection{EAAS process}
To integrate the proposed pulmonary DPM-solver with the existing anatomically aware sampling (AAS) process and further enhance model efficiency, we introduce the effective anatomically aware sampling (EAAS) method, which directly generates lung nodules. To address the imperfect blending issue between synthetic and background images, Lung-DDPM generates pulmonary regions containing lung nodules, which are then blended with extra-pulmonary areas. However, this approach is rigid and inefficient, as it only supports $128^3$ resolution due to computational constraints. To overcome this limitation, the EAAS process begins with random noise $n$, a reference CT patch $x$ cropped from a nodule-positive CT scan, and the corresponding synthetic nodule mask $m$. The input size is $64^3$, allowing the EAAS process to be executed using a lighter model. An intermediate state $x_t$ is computed by mixing the inverted $x$ and the random noise $n$ under the constraint of $m$, enabling the denoising model to dynamically blend the nodule and its surrounding pulmonary region. This inversion process follows the basic forward diffusion procedure \cite{ho2020denoising}, which gradually adds Gaussian noise to the input $x$, generating a series of intermediate states $x_1, \ldots, x_{t-1}, x_t$ with a variance $\beta_t \in (0,1)$:

\begin{equation}
\begin{aligned}
q(x_t|x_{t-1}) = \mathcal{N}(\sqrt{1-\beta_t}x_{t-1}, \beta_t\mathbf{I})
\end{aligned}
\label{eq6}
\end{equation}

Any $x_t$ at time step $t$ can be directly sampled from $x$:

\begin{equation}
\begin{aligned}
x_t=\sqrt{\bar{\alpha_t}}x+\sqrt{1-\bar{\alpha_t}}\epsilon
\end{aligned}
\label{eq7}
\end{equation}
where $\epsilon \sim \mathcal{N}(0,\mathbf{I})$, $\alpha_t=1-\beta_t$, and $\bar{\alpha_t}=\prod_{k=0}^{t}{\alpha_k}$. 

The intermediate state $x_t$ is then processed by the pulmonary DPM-solver, which performs the reverse denoising procedure. After a series of denoising steps, a clean nodule patch $x_0$ is obtained and forwarded to the next stage. The training and EAAS procedures are outlined in Algorithms \ref{alg1} and \ref{alg2}, respectively.

\begin{algorithm*}
\caption{Lung-DDPM+ training procedure}
\textbf{Input:} randomly cropped nodule patch $x_0\sim q(x_0)$ and the corresponding nodule semantic layout $m$\\
\textbf{Output:} Trained noise predictor $\hat{\epsilon}_{\theta}$
\begin{algorithmic}[1]
\Repeat 
\State $t\sim Uniform(\{ 1,...T \})$
\State $\epsilon \sim \mathcal{N}(0,\textbf{I})$
\State $x_t=\sqrt{\bar{\alpha_t}}x_0+\sqrt{1-\bar{\alpha_t}}\epsilon$
\State $\hat{x}_t=x_t\oplus m$
\State Taking a gradient step towards
\State \hskip1.5em $\nabla_\theta \mathcal{L}(\epsilon,\epsilon_\theta (\hat{x}_t, t))$
\Until{converged}
\end{algorithmic}
\label{alg1}
\end{algorithm*}

\begin{algorithm*}
\caption{Effective anatomically aware sampling (EAAS) procedure: given a pretrained noise predictor $\hat{\epsilon}_{\theta}$.}
\textbf{Input:} reference CT patch $x$, the synthetic nodule semantic layout $m$\\
\textbf{Output:} synthetic nodule CT patch $x_0$ according to $x$ and $m$
\begin{algorithmic}[1]
\State sample $n\sim \mathcal{N}(0, \textbf{I})$
\State get inverse sample $q(x_T|x)$
\State $x_T = q(x_T|x) \otimes (1-m) + n\otimes m$
\For{\textbf{all} $t$ from $T$ to 1}
    \State $x_{t-1} = P.DPM.Solver(x_t)$
\EndFor
\State \Return $x_0$
\end{algorithmic}
\label{alg2}
\end{algorithm*}

\subsubsection{Fusion}
In the final step of the sampling process, the synthetic nodule patch is fused with the original CT scan to generate the final output of Lung-DDPM+. This fusion process involves replacing the originally cropped healthy CT region with the generated CT patch containing the synthetic lung nodule. The replacement is performed based on the same coordinates used during the random cropping process in EAAS, ensuring spatial consistency. This simple yet effective strategy enables the generation of full-size CT volumes with inserted nodules while preserving the anatomical structure of the surrounding tissues.

\subsection{Model design and learning objectives}
We adopt the network architecture and learning objectives of Lung-DDPM, modifying only the model scale to accommodate a smaller input size ($64^3$).

\section{Experiments}
\label{sec4}
We conduct three evaluations in our experiments: (1) Efficiency Evaluation – This assessment focuses on analyzing and reporting the model's effectiveness. (2) Lung Nodule Segmentation Evaluation – This includes two downstream tasks: 2D and 3D lung nodule segmentation. (3) Visual Turing Test - This evaluation involves an experienced radiologist for intuitively evaluating synthetic samples' quality.

\begin{itemize} 
\item 2D Task – We assess the quality of high-resolution synthetic samples by incorporating them into the training set of a 2D lung nodule segmentation model, a common deep learning application in medical imaging.
\item 3D Task – We evaluate the model's ability to generate anatomically accurate and consistent 3D samples, emphasizing structural integrity and clinical feasibility.
\end{itemize}

\subsection{Dataset}
\subsubsection{LIDC-IDRI}

\begin{table*}[htbp] 
 \centering
 \caption{Nodule analysis for LIDC-IDRI dataset.} 
 \begin{tabular}{l  l} 
  \toprule 
  Diameter & mm \\ 
  \midrule
  Maximum  & 57.42 \\
  Minimum & 1.41  \\ 
  Average & 11.57 \\
  Median & 9.17 \\
  Standard deviation  & 7.91  \\
 \midrule 
  Nodule size & Case number (percentage) \\
 \midrule 
  Small ($<$ 6mm) & 65 (19\%) \\
  Medium (6mm $<=$ d $<$ 16 mm) & 210 (62\%) \\
  Large ($>=$ 16mm) & 66 (19\%) \\
 \bottomrule 
 \end{tabular} 
\label{table1}
\end{table*}

In our experiments, we firstly use the well-known pulmonary CT dataset LIDC-IDRI (Lung Image Database Consortium and Image Database Resource Initiative) \cite{samuel2011lung}, a widely used public benchmark for lung nodule detection, classification, and segmentation in thoracic CT scans. The dataset originally contains 1,018 anonymized low-dose chest CT scans with corresponding lung nodule annotations from four experienced radiologists, who marked and categorized nodules based on their characteristics and likelihood of malignancy.

To obtain unified nodule annotations, we apply a 50\% consensus criterion \cite{kubota2011segmentation} to manage the variability among the four radiologists. Since existing automatic segmentation methods are unstable for detecting lung nodules in contrast-enhanced CT images\cite{shaish2023diagnostic,christensen2021optimizing,ye2022deep,kihara2025impact}, we exclude 509 contrast-enhanced cases from the original dataset. Additionally, we remove 25 cases containing multiple nodules and two cases with oversized nodules ($>128$mm).

A nodule analysis was conducted on the remaining 341 cases, and the results are presented in Table \ref{table1}. Based on these results, we set the Lung-DDPM+ input size to $64^3$, aligning with the observed nodule size range. Furthermore, we categorize the cases into three groups based on nodule size distribution: small (19\%), medium (68\%), and large (19\%). When generating synthetic nodule semantic layouts, we adhere to this size distribution.

\subsubsection{NLST}
To evaluate the generalization capability of our proposed approach, we utilized the publicly available National Lung Screening Trial (NLST) dataset \cite{national2011national}, a large-scale, multi-center study comparing the effectiveness of low-dose computed tomography (LDCT) and chest X-ray (CXR) for lung cancer screening in high-risk populations. The NLST includes current or former smokers aged 55–74 years with a minimum 30 pack-year smoking history; former smokers were eligible if they had quit within the past 15 years. Notably, the original NLST dataset provides only binary malignancy labels, without explicit nodule annotations. To ensure experimental rigor and exclude benign cases with nodules, we leveraged nodule-positive cases identified by Cherezov et al. \cite{cherezov2018delta} and our previous work \cite{jiang2025benchmark}, extracting healthy slices from each case to generate synthetic nodules. In total, we selected 252 participants\footnote{One case was excluded due to resampling failure.} who underwent non-contrast CT scans. Nodule annotations were sourced from Cherezov et al. \cite{cherezov2018delta}, with all nodules manually delineated by a radiologist. Note that the NLST dataset is only used for generating synthetic samples for the lung nodule segmentation evaluation.

\subsection{Evaluation metrics}
For the efficiency evaluation, we randomly sample 100 CT images and report the average floating-point operations per second (FLOPs), average sampling time, and peak GPU memory consumption.

For the lung nodule segmentation evaluation, we perform 5-fold cross-validation with an 80\% training / 20\% test split. The proposed method and its competitors are trained using the training set from each fold and generate synthetic images based on synthetic nodule layouts derived from the training set. Finally, we train a U-Net \cite{ronneberger2015u} using a real-synthetic mixed training set and report the Dice score and Hausdorff distance (HD95) metrics, along with a 95\% confidence interval computed across the five validation folds.

In the case of the Visual Truing Test, we report the accuracy, sensitivity, and specificity for identifying a synthetic case in a real-synthetic case pair.

\subsection{Implementation details}
During data preparation, all CT images and their corresponding semantic layouts were first resampled to $1\times 1\times 1$mm. The original CT images were then randomly centrally cropped into a $64^3$ nodule patch for Lung-DDPM+ and into $16\times 512\times 512$ or $128^3$ patches for SOTA competitors in 2D and 3D lung nodule segmentation tasks, respectively. Additionally, we adopt the same training settings and hyperparameters as Lung-DDPM. For the DPM-solver parameters, we use a cosine noise schedule with 50 steps. The batch size is set to 1 across all approaches. All experiments were conducted on the HPC of Université Laval using an Nvidia Tesla A100 80GB GPU.

\subsection{Quantitative results}
\subsubsection{Efficiency evaluation}
\begin{table*}[htbp] 
\scriptsize
 \centering
 \caption{Experimental results for the efficiency evaluation. (The best evaluation score is bolded. Lower values indicate better performance. VRAM represents video random-access memory, measured in gigabytes (GB). Sampling time is measured in seconds.)} 
 \begin{tabular}{c c c c c c c} 
  \toprule 
  \makecell{Downstream task \\ (image size)} & \multicolumn{3}{c}{3D ($128^3$)} & \multicolumn{3}{c}{2D ($16\times 512\times 512$)}\\ 
  \midrule 
  Method & TFLOPs & VRAM & sampling time &  TFLOPs  & VRAM & sampling time \\ 
  \midrule 
  GEM-3D \cite{zhu2024generative} & 1.90 & 19.18 & 288.92 & 1.90  & 19.05 & 34.10\\
  Med-DDPM \cite{dorjsembe2024conditional} & 4.35 & 6.15 & 53.33 & 2.02 & 6.13 & 45.89\\
  Lung-DDPM \cite{jiang2025lungddpm} & 4.35 & 6.17 & 52.93 & 2.02 & 6.16 & 45.42\\
 \makecell{Lung-DDPM+ \\ (OURS)} & \textbf{0.54} & \textbf{0.91} & \textbf{3.79} & \textbf{0.54} & \textbf{0.91} & \textbf{3.79}\\
  \bottomrule 
 \end{tabular} 
\label{table2}
\end{table*}

Table \ref{table2} presents the efficiency evaluation results, comparing Lung-DDPM+, Lung-DDPM \cite{jiang2025lungddpm}, Med-DDPM \cite{dorjsembe2024conditional}, and GEM-3D \cite{zhu2024generative} in 3D ($128^3$) and 2D ($16 \times 512 \times 512$) lung nodule synthesis tasks. The results clearly demonstrate that Lung-DDPM+ significantly outperforms all competitors in computational efficiency, achieving the lowest TFLOPs, VRAM usage, and sampling time. Specifically, Lung-DDPM+ requires only 0.54 TFLOPs, 0.91 GB of VRAM, and 3.79 seconds per sample for both 3D and 2D synthesis, which is up to 8× fewer TFLOPs, 21× less VRAM, and 76× faster sampling time compared to Lung-DDPM and Med-DDPM. In contrast, GEM-3D exhibits the highest computational cost, with 1.90 TFLOPs, 19+ GB of VRAM usage, and significantly longer sampling times (288.92s for 3D and 34.10s for 2D), making it the least efficient method. Med-DDPM and Lung-DDPM perform similarly, requiring 4.35 TFLOPs and over 6 GB of VRAM for 3D synthesis, while their sampling times (52.93s and 53.33s, respectively) remain far from the real-time feasibility achieved by Lung-DDPM+. These findings highlight that Lung-DDPM+ maintains competitive segmentation performance while offering vastly superior efficiency, making it particularly well-suited for resource-constrained and real-time medical imaging applications.

\subsubsection{Lung nodule segmentation evaluation}
With the rapid advancement of deep learning, deep-learning-based segmentation models have evolved from 2D to 3D. While 3D segmentation models generally achieve higher accuracy and greater robustness, they require substantially more computing resources, limiting their practical deployment. In contrast, 2D segmentation models have lower precision due to the lack of 3D nodule texture and spatial context, but they benefit from a lighter model design and improved computational efficiency. Given that both 3D and 2D models are widely adopted in clinical and research applications, and to comprehensively evaluate the quality of synthetic samples generated by the proposed method and ensure comparability with other SOTA generative models, we design two sub-tasks: lung nodule segmentation evaluation using 3D and 2D CT images.

Both 3D and 2D lung nodule segmentation evaluations follow a similar procedure:
\begin{itemize}
    \item[1.] Model Training – The proposed method and other competitive models are trained using the training set from each cross-validation fold, resulting in five different pretrained models.
    \item[2.] Synthetic Data Generation – Synthetic lung nodule semantic layouts are generated based on the cases in the cross-validation training sets and used as conditions for sampling synthetic CT images.
    \item[3.] Training Set Augmentation – Synthetic CT images are combined with the cross-validation training sets at varying synthetic-to-real ratios (100\% to 1000\%) to create a mixed training set for training a U-Net \cite{ronneberger2015u} segmentation model.
    \item[4.] Evaluation – Segmentation metrics are reported on the cross-validation test sets.
\end{itemize}

Note that the 3D U-Net model is trained using $128^3$ volumes, while the 2D U-Net model is trained on $512\times 512$ slices randomly selected from synthetic $16\times 512\times 512$ volumes.

\begin{table*}[htbp] 
\scriptsize
 \centering
 \caption{Experimental results for the lung nodule segmentation evaluation on the LIDC-IDRI dataset. (The best evaluation score is bolded. $\uparrow$ indicates that a higher value is better, while $\downarrow$ denotes that a lower value is better. Syn. indicates synthetic samples.)} 
 \begin{tabular}{c c c c c c} 
  \toprule 
  Evaluation &  & \multicolumn{2}{c}{3D} & \multicolumn{2}{c}{2D}\\ 
  \midrule 
  Method & \makecell{Real + \\  N\% Syn.} & Dice ($\uparrow$) & HD95 ($\downarrow$) & Dice ($\uparrow$) & HD95 ($\downarrow$) \\ 
  \midrule 
  & Real only & 0.4798±0.0465 & 55.6362±9.9893 & 0.0195±0.0077 & 299.9889±16.8813 \\
  \midrule 
  RandRotate & - & 0.4522±0.0662 & 68.1325±5.0597 & 0.0149±0.0041 & 317.7029±23.0801 \\
  RandFlip & - & 0.4619±0.0318 & 65.9160±5.4337 & 0.0156±0.0073 & 307.4255±17.9159 \\
  RandAffine & - & 0.4600±0.0395 & 65.2702±4.9102 & 0.0141±0.0054 & \textbf{302.1400±17.5836} \\
  RandGaussianNoise & - & \textbf{0.4675±0.0653} & \textbf{62.0637±3.2869} & \textbf{0.0194±0.0085} & 306.3446±10.8929 \\
  Ensemble & - & 0.4509±0.0695 & 62.3101±7.1750 & 0.0169±0.0058 & 304.2819±12.8653 \\
  
  \midrule 
  \multirow{9}{4em}{\makecell{Lung-DDPM+ \\ (OURS)}} & +100\% & 0.5041±0.0303 & 47.9249±2.9606 & 0.2898±0.0987 & 187.7061±44.9633 \\
  & +200\% & 0.5355±0.0666 & 36.3203±8.2924 & 0.4174±0.0698 & 147.5545±22.3728 \\
  & +300\% & 0.5380±0.0518 & 34.5296±3.1890 & 0.4191±0.0450 & 141.8068±30.9338 \\
  & +400\% & 0.5513±0.0571 & 28.4985±4.3345 & 0.4165±0.0499 & 134.9405±16.3222 \\
  & +500\% & 0.5486±0.0495 & 28.4139±6.2823 & \textbf{0.4460±0.0477} & 135.2457±18.3273 \\
  & +600\% & 0.5482±0.0640 & 26.9699±3.7543 & 0.4389±0.0506 & 123.4654±7.7184 \\
  & +700\% & 0.5597±0.0383 & 26.1724±2.1996 & 0.4435±0.0305 & \textbf{114.7386±12.7012} \\
  & +800\% & 0.5522±0.0600 & 25.1108±4.3400 & 0.4178±0.0487 & 131.9875±28.3479 \\
  & +900\% & \textbf{0.5626±0.0467} & \textbf{24.6928±3.2572} & 0.4241±0.0560 & 135.7615±33.6063 \\
  & +1000\% & 0.5498±0.0473 & 24.9184±3.4168 & 0.4150±0.0519 & 121.3677±23.6122 \\
  \midrule 
  \multirow{9}{4em}{\makecell{Lung-DDPM \\ \cite{jiang2025lungddpm}}} & +100\% & 0.5083±0.0501 & 45.7542±3.2250 & 0.2625±0.0617 & 174.8173±49.8686 \\
  & +200\% & 0.5222±0.0565 & 39.9508±2.5507 & 0.4502±0.0418 & 119.7638±16.4344 \\
  & +300\% & 0.5373±0.0661 & 36.3010±6.7457 & 0.4491±0.0487 & 131.8097±11.1773 \\
  & +400\% & 0.5307±0.0480 & 35.3176±7.0974 & 0.4639±0.0532 & 129.7896±21.1006 \\
  & +500\% & 0.5447±0.0676 & 30.3273±4.9410 & \textbf{0.4670±0.0427} & \textbf{118.9972±23.2433} \\
  & +600\% & \textbf{0.5484±0.0743} & 33.2802±4.6414 & 0.4613±0.0301 & 135.8870±21.0061 \\
  & +700\% & 0.5393±0.0482 & 31.2259±2.3348 & 0.4503±0.0616 & 129.3129±22.1311 \\
  & +800\% & 0.5393±0.0373 & 31.8385±5.2278 & 0.4408±0.0730 & 134.6253±39.1078 \\
  & +900\% & 0.5398±0.0453 & 32.5331±5.2594 & 0.4614±0.0641 & 127.6917±21.9044 \\
  & +1000\% & 0.5457±0.0347 & \textbf{28.9722±4.1263} & 0.4464±0.0614 & 135.6789±25.4054 \\
  \midrule 
  \multirow{9}{4em}{\makecell{GEM-3D \\ \cite{zhu2024generative}}} & +100\% & 0.4718±0.0450 & 53.2418±4.1788 & 0.1074±0.0974 & 246.5404±45.7044 \\
  & +200\% & \textbf{0.4720±0.0563} & 45.3387±6.0707 & 0.3829±0.0484 & 149.5858±39.3764 \\
  & +300\% & 0.4617±0.0484 & 39.0576±6.0297 & 0.3458±0.0798 & 167.6902±42.3420 \\
  & +400\% & 0.4610±0.0692 & 41.7313±3.8246 & \textbf{0.4007±0.0668} & 153.6798±23.8837 \\
  & +500\% & 0.4426±0.0188 & 37.4128±2.4877 & 0.3924±0.0540 & \textbf{134.3399±30.2846} \\
  & +600\% & 0.4508±0.0497 & 37.6485±2.4520 & 0.2980±0.2074 & 173.4189±72.8671\\
  & +700\% & 0.4203±0.0582 & \textbf{35.9151±7.3135} & 0.3185±0.0512 & 170.6918±28.2373 \\
  & +800\% & 0.4181±0.0680 & 39.1341±2.1743 & 0.2593±0.1007 & 174.9276±19.0021 \\
  & +900\% & 0.4228±0.0376 & 36.5975±2.3046 & 0.3862±0.0794 & 156.2082±45.3715 \\
  & +1000\% & 0.3959±0.0623 & 36.8016±4.3721 & 0.2512±0.1117 & 173.0495±14.2321 \\
  \midrule 
  \multirow{9}{4em}{\makecell{Med-DDPM \\ \cite{dorjsembe2024conditional}}} & +100\% & 0.5135±0.0588 & 44.6145±6.1502 & 0.2868±0.0741 & 179.7118±30.6357 \\
  & +200\% & 0.5271±0.0575 & 40.7760±2.5872 & 0.4453±0.0385 & 125.0936±25.0093 \\
  & +300\% & 0.5409±0.0779 & 36.0338±8.0448 & 0.4328±0.0200 & 138.8558±20.2925 \\
  & +400\% & 0.5582±0.0544 & 30.6473±6.0894 & 0.4558±0.0238 & 121.4687±22.2562 \\
  & +500\% & 0.5539±0.0675 & 29.8979±7.6026 & \textbf{0.4667±0.0465} & 124.0373±18.8632 \\
  & +600\% & 0.5479±0.0490 & 31.1199±6.8460 & 0.4620±0.0445 & 114.5856±22.3203 \\
  & +700\% & 0.5591±0.0389 & 29.2162±3.1307 & 0.4609±0.0415 & 113.3781±24.2374 \\
  & +800\% & 0.5600±0.0638 & 30.4333±5.8696 & 0.4623±0.0455 & 110.8988±14.4987 \\
  & +900\% & 0.5571±0.0718 & 28.1161±4.2287 & 0.4644±0.0351 & \textbf{107.4805±6.1294} \\
  & +1000\% & \textbf{0.5640±0.0612} & \textbf{26.8428±7.2266} & 0.4380±0.0578 & 123.8135±25.3507 \\
  \bottomrule 
 \end{tabular} 
\label{table3}
\end{table*}

Table \ref{table3} presents the lung nodule segmentation evaluation results, comparing traditional data augmentation methods, Lung-DDPM+, Lung-DDPM \cite{jiang2025lungddpm}, GEM-3D \cite{zhu2024generative}, and Med-DDPM \cite{dorjsembe2024conditional} across 2D and 3D segmentation tasks. The results demonstrate that Lung-DDPM+ achieves comparable segmentation performance to other SOTA models while maintaining superior computational efficiency. In 3D lung nodule segmentation, Med-DDPM achieves the highest Dice score (0.5640 at 1000\% synthetic data) and the lowest HD95 (26.8428 at 1000\%), while Lung-DDPM+ closely follows with a Dice score of 0.5626 at 900\% and an HD95 of 24.6928, which is lower than Med-DDPM's best HD95 score, suggesting that Lung-DDPM+ maintains strong anatomical precision. In 2D segmentation, Lung-DDPM (0.4670 at 500\%) and Lung-DDPM+ (0.4460 at 500\%) achieve comparable Dice scores, while Med-DDPM (HD95 = 107.4805 at 900\%) maintains the lowest boundary error. Notably, Lung-DDPM+ consistently outperforms GEM-3D, which struggles to generate meaningful segmentation results, with significantly lower Dice scores (max 0.4720 at 200\%) and higher HD95 values across all settings. These findings highlight that Lung-DDPM+ matches or exceeds the performance of other generative models while achieving better sampling efficiency, requiring less computational overhead than Lung-DDPM and Med-DDPM while still producing high-quality lung nodules with clear anatomical structures. Furthermore, Lung-DDPM+ effectively leverages synthetic data, demonstrating that a balanced synthetic-to-real data ratio (500\%–900\%) significantly boosts segmentation performance, particularly in 3D segmentation tasks, where Lung-DDPM+ exhibits a clear advantage in maintaining detailed anatomical structures. These results confirm that Lung-DDPM+ provides an optimal balance between efficiency and segmentation accuracy, making it a compelling choice for medical imaging applications that require both high-quality synthetic data generation and real-time feasibility.

Table 3 also shows that traditional augmentation methods do not bring meaningful improvements in either 2D or 3D lung nodule segmentation. For the 3D task, Dice scores for all traditional augmentation methods remain close to or below the baseline (real-only) result of 0.4798, with no method achieving better accuracy. Similarly, in the 2D task, Dice scores remain extremely low (ranging from 0.0141 to 0.0194), suggesting no benefit from these augmentations. In contrast, the proposed Lung-DDPM+ significantly boosts segmentation performance, reaching Dice scores as high as 0.5626 in 3D and 0.4460 in 2D. These results indicate that while traditional augmentations are commonly used, they are ineffective in our setting, and model-driven data augmentation using Lung-DDPM+ offers a more powerful and reliable strategy for enhancing downstream segmentation performance.

\begin{figure}[!ht]
\centering
\includegraphics[width=14cm]{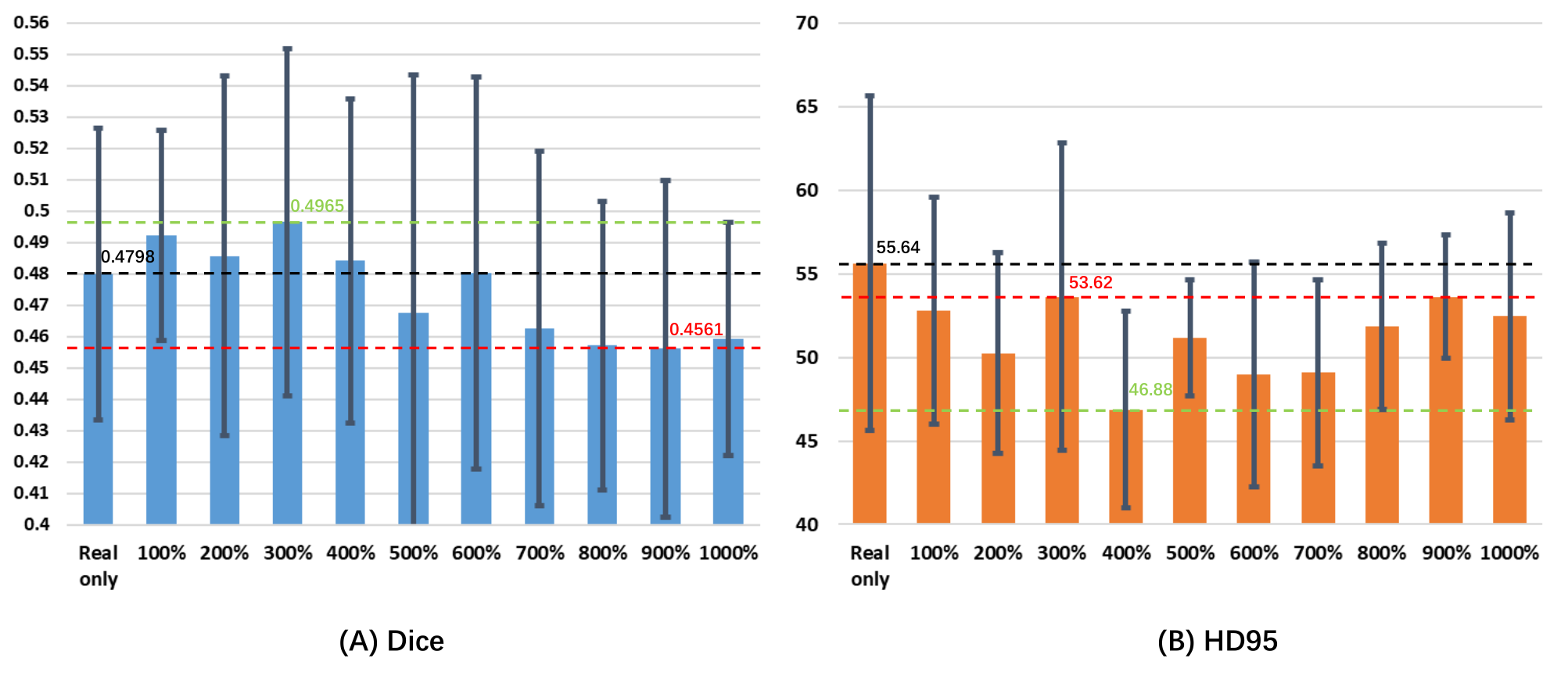}
\caption{Experimental results for 3D lung nodule segmentation on the NLST dataset. Subfigures (A) and (B) display the Dice and HD95 metrics, respectively. The green dashed line indicates the best performance, the red dashed line marks the worst, and the black dashed line represents the baseline metric. Note that 100\%-1000\% indicates the percentage of the synthetic NLST cases that are included in the training set.}
\label{fig4} 
\end{figure}

Figure \ref{fig4} illustrates the 3D lung nodule segmentation performance of Lung-DDPM+ on the NLST dataset under varying synthetic-to-real data ratios. In subfigure (A), Dice performance slightly improves with the inclusion of synthetic data, peaking at 0.4965 with 300\% synthetic cases, but gradually declines thereafter, reaching a minimum of 0.4561 at 900\%. In subfigure (B), all HD95 values are better than the baseline of 55.64, indicating improved boundary precision with synthetic data. The best HD95 of 46.88 is achieved at 400\%, showing that even moderate amounts of synthetic data can enhance segmentation accuracy. However, when compared to Table \ref{table3}, which presents results on the LIDC-IDRI dataset, the improvements on NLST are relatively modest. For example, Lung-DDPM+ reaches a higher Dice score of 0.5626 and a substantially lower HD95 of 24.6928 on LIDC-IDRI. This discrepancy suggests that differences in scanning protocols, hardware, or population characteristics between the two datasets may limit the generalizability or impact of synthetic augmentation, underscoring the importance of dataset-specific considerations when applying generative models in medical imaging.

\begin{figure}[!ht]
\centering
\includegraphics[width=14cm]{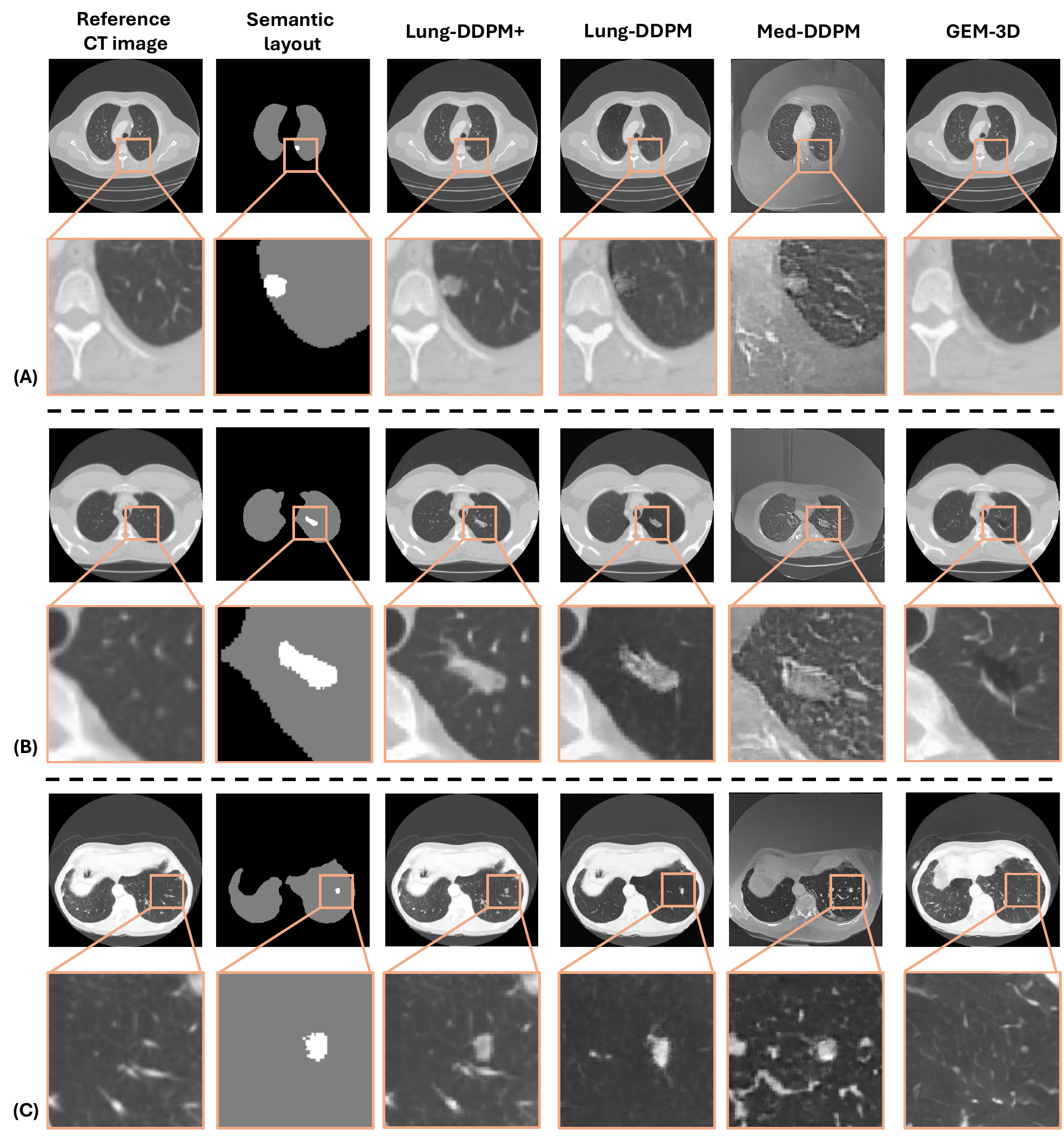}
\caption{Demonstration of synthetic samples generated by Lung-DDPM+ and other SOTA competitors. Gray areas in the semantic layouts represent pulmonary regions, while white areas indicate lung nodule regions.}
\label{fig5} 
\end{figure}

\subsection{Qualitative results}
In this subsection, we evaluate the quality of synthetic samples by visually interpreting them by an experienced radiologist and demonstrating them by comparing them with those generated by other SOTA methods. 

The Visual Turing Test was conducted on 61 randomly selected real-synthetic case pairs, comprising 61 real CT scans and 61 synthetic CT scans. Each pair included a real case, the corresponding lung nodule segmentation, and synthetic cases generated based on the segmentation. The radiologist was asked to identify which case was fake in each pair. Following the methodologies outlined in \cite{chen2024towards} and \cite{hu2023label}, all pairs were provided as 2D animations to effectively represent the 3D CT volume while focusing exclusively on the nodule-positive slices. The evaluation results indicate that the radiologist achieved an overall accuracy of 0.6393, with a sensitivity of 0.7647 and a specificity of 0.4815. These results demonstrate that while the radiologist exhibited a relatively high ability to correctly identify synthetic cases when a fake was present (high sensitivity), the specificity was notably lower, indicating challenges in distinguishing real cases when presented with the synthetic alternatives. The lower specificity suggests that certain characteristics of the synthetic images closely mimicked real CT scans, potentially leading to false-positive identifications. This discrepancy between sensitivity and specificity underscores the realism of the synthetic data, particularly in replicating the nuanced features of real lung nodules, and highlights the potential of the proposed Lung-DDPM+ in contributing to robust synthetic datasets for different downstream thoracic imaging applications.

In Figure \ref{fig5}, we present three cases, each showing the reference CT image, semantic layout, and synthetic samples generated by Lung-DDPM+, Lung-DDPM, Med-DDPM, and GEM-3D. Overall, GEM-3D failed to generate a meaningful lung nodule in all cases. Med-DDPM produced synthetic samples that were generally noisy and exhibited unnatural artifacts in both extra-pulmonary and pulmonary regions. Additionally, Med-DDPM failed to generate coherent structures in the extra-pulmonary region due to the absence of semantic layout guidance. Lung-DDPM generated low-quality lung nodules with noise and artifacts. In contrast, the proposed Lung-DDPM+ produced high-quality lung nodules with clear boundaries, while preserving anatomical details beyond the synthetic nodule regions.

\section{Discussion}
\label{sec5}
In this section, we analyze the experimental results and examine how the proposed Lung-DDPM+ addresses existing challenges while maintaining performance comparable to other SOTA generative models.

The efficiency of generative models has been widely studied and recognized as a critical factor for practical deployment \cite{liu2024deep}. However, improving efficiency remains challenging, particularly for emerging approaches with complex architectures and diverse applications. Various efforts have been made to enhance computational efficiency across SOTA generative models, including generative adversarial networks (GANs) \cite{tao2022df}, variational autoencoders (VAEs) \cite{rybkin2021simple}, and diffusion models \cite{ma2024efficient}. Despite these advancements, most efficiency improvements have been tailored to general visual tasks, rather than medical imaging applications. In this paper, we introduce Lung-DDPM+, a novel diffusion-based model for efficient lung nodule CT image synthesis, achieving significant efficiency gains over SOTA generative models. As shown in Table \ref{table2}, Lung-DDPM+ drastically reduces computational overhead, requiring up to 8× fewer TFLOPs, 21× less VRAM, and 76× shorter sampling time compared to Lung-DDPM and Med-DDPM. Given its real-time generation capability, Lung-DDPM+ is particularly well-suited for cost-sensitive and edge-computing scenarios, enabling the broader adoption of high-quality synthetic medical imaging data.

In addition to efficiency, anatomical precision is a crucial aspect of synthetic medical imaging. Our previous work, Lung-DDPM \cite{jiang2025lungddpm}, introduced a semantic layout-guided diffusion model to generate pulmonary regions, which were then blended with extra-pulmonary areas using an anatomically aware sampling process. While Lung-DDPM partially addressed the anatomical imprecision issue, it still struggled to reconstruct fine pulmonary structures, such as vasculature and the bronchial tree, due to their small size and low contrast. Instead of synthesizing an entire pulmonary region, Lung-DDPM+ exclusively generates nodule regions, which are seamlessly integrated into surrounding tissues. This novel approach allows the model to focus on nodule texture while preserving anatomical accuracy. As demonstrated in Figure \ref{fig5}, this targeted synthesis strategy significantly improves structural consistency and generation stability compared to other SOTA methods.

Beyond efficiency and anatomical accuracy, synthetic lung nodule CT images play a vital role in downstream applications, such as lung cancer risk prediction \cite{jiang2025benchmark}, tumor segmentation \cite{jiang2018tumor}, and lung cancer survival prediction \cite{lai2020overall}. To assess synthetic sample quality, we evaluate Lung-DDPM+ on 2D and 3D lung nodule segmentation tasks, using Dice scores and HD95 metrics as performance indicators. As shown in Table \ref{table3}, synthetic samples enhance segmentation model performance across both 2D and 3D tasks. However, achieving optimal performance requires a significant proportion of synthetic data—typically above 500\%—and each model exhibits a distinct sweet spot for the ideal synthetic-to-real data ratio. Notably, 3D segmentation models demand more synthetic data than 2D models to attain meaningful improvements, emphasizing the need for carefully designed data augmentation strategies. Additionally, we observe a marginal effect, where the performance gains saturate as more synthetic data is introduced. This underscores the importance of balancing synthetic data volume, generation cost, and existing dataset scale to maximize segmentation accuracy while maintaining computational efficiency.

While our experiments were conducted using the standard U-Net architecture to maintain methodological clarity and enable controlled comparisons across generative models, we acknowledge that more advanced segmentation frameworks have emerged in recent years. Architectures such as nnU-Net \cite{isensee2021nnu}, which incorporates automated pipeline optimization and extensive built-in augmentation strategies, represent the current state-of-the-art in medical image segmentation. Future work could extend our augmentation strategy to such advanced and automated segmentation pipelines to evaluate whether the benefits of synthetic data generation persist or even amplify when combined with more sophisticated architectures. Given that nnU-Net already includes comprehensive traditional augmentation methods, investigating how model-based synthetic data augmentation complements or enhances these existing strategies would be particularly valuable. Such studies would help establish whether our approach provides additive benefits in highly optimized segmentation frameworks and could further validate the generalizability of Lung-DDPM+ across diverse clinical applications and architectural choices.

Overall, these results confirm that Lung-DDPM+ provides a well-balanced solution, offering state-of-the-art segmentation performance with significantly improved efficiency. By integrating anatomically aware synthesis and highly efficient sampling, Lung-DDPM+ not only matches or surpasses competing diffusion models in segmentation tasks but also achieves a level of efficiency that makes it practical for real-time medical imaging applications. Future work will focus on extending this framework to broader tumor types, organ structures, and imaging modalities, further expanding the applicability of generative models in clinical and research environments.

\section{Conclusion and future study}
\label{sec6}
In this paper, we propose Lung-DDPM+, a novel CT image synthesis method capable of generating high-quality nodule-positive CT images efficiently. The proposed pulmonary anatomically aware sampling process enables Lung-DDPM+ to synthesize realistic lung nodules from healthy CT slices while reducing computational complexity and hardware requirements compared to its predecessor. Additionally, Lung-DDPM+ preserves detailed and anatomically accurate structures throughout the sampling process. Experimental results demonstrate that the proposed method outperforms SOTA competitors in sampling efficiency and downstream lung nodule segmentation tasks, highlighting its potential in various medical imaging applications. 

While our experiments employed the standard U-Net architecture for methodological clarity and controlled comparison, future work could extend our augmentation strategy to more advanced segmentation frameworks such as nnU-Net. Investigating how model-based synthetic data complements the sophisticated built-in augmentation strategies of such pipelines would help establish the broader applicability of Lung-DDPM+ across diverse architectural choices and clinical applications.

Although our nodule semantic layout generation uses ellipsoidal approximations, our results demonstrate that this simplified approach achieves substantial segmentation improvements, suggesting the diffusion model generates diverse nodule textures beyond initial geometric constraints. Future work could investigate whether incorporating more sophisticated morphological models for spiculated, ground-glass, or lobulated presentations would further enhance synthetic data quality and downstream performance.

Additionally, we will explore how to extend this method to a broader range of tumors, organs, and medical imaging modalities to enhance the deployability of generative models in clinical and research settings.

\section{Acknowledgments}
\label{sec7}
Venkata SK. Manem holds a salary support award from the IVADO and supported by the CHU de Quebec start-up funds.


 \bibliographystyle{elsarticle-num-names} 
 \bibliography{main}






\end{document}